\documentclass{article}
\usepackage{ijcai23}

\usepackage{times}
\usepackage{soul}
\usepackage{url}

\usepackage[hidelinks]{hyperref}
\usepackage[utf8]{inputenc}
\usepackage[small]{caption}
\usepackage{graphicx}
\usepackage{amsmath}
\usepackage{amsthm}
\usepackage{booktabs}
\usepackage{algorithm}
\usepackage{hyperref}

\usepackage[switch]{lineno}
\usepackage{hyperref}
\usepackage{url}

\usepackage{multicol}

\usepackage{paralist}
\usepackage{float,bbold}
\usepackage{dsfont}

\usepackage{algpseudocode}

\usepackage{todonotes}
\usepackage{ verbatim,subcaption}

\newtheorem{example}{Example}
\newtheorem{theorem}{Theorem}

\title{Counterfactual Explanation via Search in Gaussian Mixture Distributed Latent Space}

\author{
Xuan Zhao$^1$
\and
Klaus Broelemann$^1$\and
Gjergji Kasneci$^{2}$
\affiliations
$^1$SCHUFA Holding AG\\
$^2$Technical University of Munich\\
\emails
xuan.zhao@schufa.de,
klaus.broelemann@schufa.de,
gjergji.kasneci@tum.de
}

\begin{document}

\maketitle

\begin{abstract}
Counterfactual Explanations (CEs) are an important tool in Algorithmic Recourse for addressing two questions: 1. What are the crucial factors that led to an automated prediction/decision? 2. How can these factors be changed to achieve a more favorable outcome from a user's perspective? Thus, guiding the user's interaction with AI systems by proposing easy-to-understand explanations and easy-to-attain feasible changes is essential for the trustworthy adoption and long-term acceptance of AI systems. In the literature, various methods have been proposed to generate CEs, and different quality measures have been suggested to evaluate these methods. However, the generation of CEs is usually computationally expensive, and the resulting suggestions are unrealistic and thus non-actionable. In this paper, we introduce a new method to generate CEs for a pre-trained binary classifier by first shaping the latent space of an autoencoder to be a mixture of Gaussian distributions. CEs are then generated in latent space by linear interpolation between the query sample and the centroid of the target class. We show that our method maintains the characteristics of the input sample during the counterfactual search. In various experiments, we show that the proposed method is competitive based on different quality measures on image and tabular datasets -- efficiently returns results that are closer to the original data manifold compared to three state-of-the-art methods, which are essential for realistic high-dimensional machine learning applications.
\end{abstract}

\section{Introduction}\label{sec:intro}

In contemporary society, the widespread adoption of machine learning models is evident, with a growing emphasis on the need for transparency, particularly in critical domains such as healthcare, finance, and employment. The ability to comprehend the behavior of these models is essential for instilling trust in decision-making systems rooted in machine learning. Various methodologies have been devised by researchers to elucidate the connections between a model's input and output. While simpler models like logistic regression, decision trees, and rule fit algorithms allow for direct interpretation, more intricate models necessitate analysis through the lens of simpler surrogate models. 
An additional avenue of investigation focuses on addressing a fundamental query within a binary classification context: ``How can one modify the input to generate a prediction favoring the preferred class over the unfavored one?" This exploration, known as counterfactual analysis, involves examining outcomes in alternative yet similar non-occurring scenarios \cite{menzies2020}. When working with a pre-trained and fixed model, the sole method for altering the model's output is by modifying the input. Counterfactual explanations (CEs) serve as prescriptive recommendations regarding the features of the queried sample (i.e., input) that need adjustment (along with the magnitude of the required changes) to attain the desired result. It is essential for these changes to be minimal (associated with low costs) and feasible (realistic). With clear semantic implications and a common logical foundation, CEs are generally comprehensible for end-users~\cite{fernandez2020}. In fact, in the literature, the most direct application of CEs often involves providing guidance to end-users.

However, CEs have their downsides. High-dimensional input spaces lead to high-dimensional CEs with limited utility for the potential users of the explanation (less intuitive and less feasible). 
In addition, searching through the high-dimensional space for CEs is computationally expensive. Other than the problem caused by the dimensionality, without proper constraints, it is possible to generate out-of-sample CEs close to the original data distribution. Out-of-sample CEs can result in explanations/suggestions that are not feasible (i.e., unlikely to be achieved since they do not correspond to the training data distribution). Adversarial samples \cite{goodfellow2015} which resemble the original sample but have changed imperceptibly to fool the classifier, might be a good example to illustrate this situation even though they are designed for a different purpose (deceive human perception and pre-trained classifiers). We show in Section \ref{sec:eva} that a search in the original space for CEs might cause this situation. An feasible CE should stay close to the data manifold and suggest meaningful (i.e., realistic and easily achievable) changes to a query sample.   
Various requirements are proposed in the literature for generating useful CEs: feasible, sparse, valid, proximate, and computationally efficient to generate~\cite{verma2020}. It is also clear that there might be a trade-off between these requirements. 

We introduce a method for generating CEs by using interpolation within the latent representations of the input data to achieve the requirements mentioned above. We perform experiments with an image dataset -- MNIST and two tabular datasets -- the Adult income and Lending Club loan default. Our main contributions are: (1) a model agnostic framework for finding feasible CEs that are prominent in scalability in data with a low computational expense, (2) a novel strategy for manipulating the latent space for the counterfactual search, (3) a comparison of methods across the image and tabular datasets. 

The remainder of this paper is structured as follows: First, we briefly review the existing methods of counterfactual explanation (Section \ref{sec:back}). Then, in Section \ref{sec:method}, we propose a CE method via autoencoder-aided search in Gaussian Mixture (GM) distributed latent space. In Section \ref{sec:eva}, we present the evaluation results of our method against three state-of-the-art methods. Finally, Section \ref{sec:con} concludes the paper.

\section{Background and Related Work}
\label{sec:back}
\subsection{Generating Counterfactuals by Perturbing the Original Input Space}
A substantial portion of the existing literature focuses on generating counterfactuals by perturbing the input feature space. In the realm of inverse classification \cite{lash2017}, sparsity is preserved by categorizing features into immutable and mutable ones, subjecting the latter to budgetary constraints. An alternative approach, proposed by \cite{laugel2017}, employs the growing spheres method for sampling to traverse the input space and identify Counterfactual Explanations (CEs). Utilizing gradient descent in the input space, \cite{dhurandhar2018} seeks contrastive explanations, categorizing them into pertinent positives and pertinent negatives, and incorporating an autoencoder loss to ensure explanations remain within the sample.

Further enhancements to gradient descent methods involve the introduction of prototypes, guiding the descent towards the average value of the target class by averaging representations in the latent space from the training sets \cite{vanlooveren2020}. GRACE \cite{le2020} is specifically tailored for neural networks on tabular data, combining contrastive explanations with interventions through constrained gradient descent. It adds an extra loss, measuring information gain, to maintain sparse resulting explanations. Many of the aforementioned methods already incorporate generative models to stay within the sample. The Explainable Artificial Intelligence (XAI) criteria for CEs are integrated into these methods, either as constraints or as the backbone of the loss design, which may pose challenges for optimization, particularly in high-dimensional input scenarios. 

\subsection{Generating Counterfactuals by Perturbing the Latent Space}
Methods that perturb input features lacking proper regularization may yield unconvincing and impractical counterfactuals, bearing a resemblance to adversarial samples \cite{goyal2019b}. To tackle this issue, employing latent space perturbation methods proves promising, as they integrate generative and probabilistic models into the algorithm design, ensuring that counterfactuals align with a high probability under the data distribution $p(X)$. An illustrative instance is StylEx \cite{lang2021}, wherein the classifier is incorporated into the generative model, manipulating the latent space to facilitate the visualization of counterfactual searches.

ExplainGAN \cite{samangouei2018} offers another approach, focusing on generating counterfactual explanations for images. This method involves training multiple autoencoders and leveraging signals from classifiers and discriminators to inform the learned representations. Counterfactuals are then generated through a growing sphere search in the latent space of a conditional variational autoencoder \cite{pawelczyk2020}. In a different vein, \cite{GuyomardFBG21} creates counterfactuals using class prototypes derived from a supervised auto-encoder, a unique aspect being the use of supervised auto-encoders for obtaining these prototypes.

For gradient descent in the latent space of a variational autoencoder, regularization terms are applied to enhance the process \cite{balasubramanian2021}. Despite their efficacy, these methods may incur computational expenses. An alternative, Sharpshooter \cite{barr2021}, employs linear interpolation in the latent space with the assistance of two distinct autoencoders, presenting a less costly approach. However, it does not guarantee the preservation of label-irrelevant features in the counterfactual search.

\section{Our method}
\label{sec:method}





\subsection{Problem Setting}
Our method requires access to the dataset and a previously trained classifier that we want to explain with counterfactual explanations. We only consider a binary classification in this paper. In the problem setting of counterfactual explanation, there is usually a base class where instances belong that class wants to seek feasible counterfactual explanation toward the target class. The dataset is composed $\mathcal{D} = (x_i,y_i)_{i=1}^K$ where we can split $x$ into two classes where $x_i$ belongs to the base class has $\hat{y_i}= 0$ and $x_{i}$ belongs to the target class has $\hat{y_i} = 1$ under the classifier $\hat{y} = f(x)$. $x_q$ is a query sample with an output 0 from the classifier $f$. 1 is the desired target output. Hence, CEs are needed for the query sample $x_q$ under the classifier $f$.



\begin{figure*}[bt] 
    \centering
    \includegraphics[width=13cm]{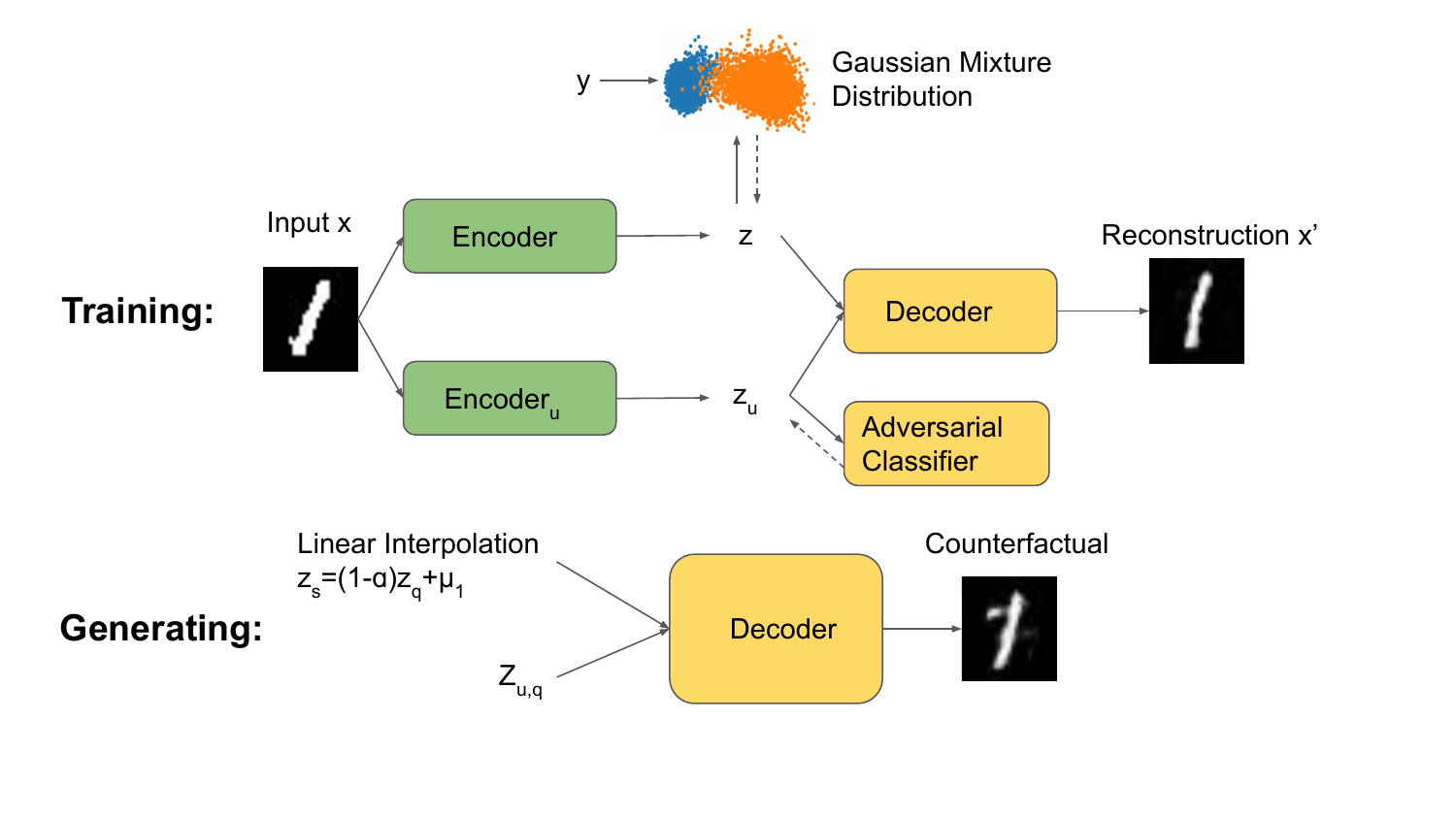}
    \caption{Diagram of the data flow in the algorithm. During Training: an input sample $x$ from the training dataset is
embedded in a latent space with a Gaussian mixture distribution by the Encoder. $x$ is also embedded into a latent space to get $z_u$ learned by the Encoder\textsubscript{u} with an Adversarial Classifier that helps remove the information related to the classification label $\hat{y}$ of $x$ under the pre-trained classifier $f$. $z$ and $z_u$ are then combined to get reconstruction $x'$. During Generating: $x_q$ is a certain query sample we need to explain with CE. $z_q$ is Encoder($x_q$) and $\mu_1$ is the mean of target class in latent space $z$. A linear search is performed in the latent space of $z$ to get a searched potential $z_s$ ($a$ is a parameter that controls the interpolation). Then $z_s$ and $z_{u,q}$ are projected back to the original space by the Decoder until the decision boundary of the classifier is crossed. }
    \label{fig:1}
\end{figure*}

\subsection{Desiderata for CE Search}
\label{Desiderata}

Our approach is driven by the following main desiderata:

\textbf{Desideratum 1}
The exploration process must preserve intrinsic and label-irrelevant attributes of the input query. This implies that the attainment of the target label should rely on the inherent properties of the input and demand minimal effort.

\textbf{Desideratum 2} The generating The generation step should produce counterfactual explanations (CEs) that are notably realistic, ensuring feasibility.

\textbf{Desideratum 3} 
Efficient retrieval of CEs is essential, especially in high-dimensional scenarios, to guarantee practical applicability in real-life situations.

\subsection{Learning Components}
We design a CE generation method based on two main steps: a Training Step in which we use the same training dataset used in training the classifier to learn an autoencoder with its latent space shaped by enforcing a Gaussian-mixture distribution on the embeddings and a Generating Step in which we use interpolation in the latent space to find a relatively less computational expensive counterfactual explanation. The algorithm flow is shown in Figure \ref{fig:1}. It only requires access to the training dataset and prediction of the pre-trained classifier $f$ that we aim to explain through appropriate CEs. For our Training Step, we form a new training dataset $\mathcal{D}_t = (x_i,\hat{y}_i)_{i=1}^K$ by replacing original $y$ in $\mathcal{D}$ with $\hat{y}$.

In the following part of this subsection, we will describe each component of the Training Step in Figure \ref{fig:1}. The details of Generating Step are described in Section \ref{sec:algorithm}.

\textbf{Label Relevant Branch: Gaussian Mixture Distribution} 
Gaussian Mixture models are usually used in a unsupervised way, but in our usage, it is supervised by the classification label produced by the pre-trained classifier. Our goal for generating a CE is to cross the decision boundary of the pre-trained classifier by taking advantage of Gaussian mixture distribution in the latent space, in which case, we could generate a CE without having a process of optimization for each query sample. Intuitively, in the latent space, data points with the same class label should cluster closer to each other. Inspired by GM loss \cite{wan2018}, we could use a classification constraint and likelihood constraints to 'push' the latent space to a Gaussian mixture distribution for further manipulation.  
After proper training, we could `force' the extracted embedding $z$ on the training set following a Gaussian mixture distribution expressed in Equation \ref{eq:z}, in which $\mu_{c}$ and $\sigma_{c}$ are the mean and covariance of class c in the latent space, and $p(c)$ is the prior probability of class $c$. 
 In the binary classification setting, $c\in [0,1]$.

\begin{align}
    \label{eq:z}
        p(z) & =\sum_{c}p(z \mid c)p(c)=\sum_{c}\mathcal{N}(z;\mu_{c},\sigma_{c})p(c)
\end{align}



If the latent space follows a Gaussian mixture distribution, the conditional probability distribution of a latent embedding $z$ given its class label c can be expressed in Equation \ref{eq:zc}. The corresponding posterior probability distribution can be expressed in Equation \ref{eq:cz}.
\begin{align}
    \label{eq:zc}
p(z \mid c)=\mathcal{N}(z;\mu_{c},\sigma_{c})p(c)
        \\
\label{eq:cz}
p(c \mid z) = \frac{\mathcal{N}( z;\mu_{c},\Sigma_{c})p(\mu_{c})}{\sum_{c=1}^{C}\mathcal{N}(z;\mu_{c},\sigma_{c})}
\end{align}
A \textbf{classification loss} $\mathcal{L}_{cls}$ is then calculated as the cross-entropy between the posterior probability distribution and the class label as is shown in Equation \ref{eq:cls}.

\begin{equation} \label{eq:cls}
\mathcal{L}_{cls} = -\frac{1}{N}\sum_{i=1}^{N}\text{log}\frac{\mathcal{N}( z_i;\mu_{\hat{y}},\Sigma_{\hat{y}}) p(\mu_{c=\hat{y}})}{\sum_{c=1}^{C}\mathcal{N}(z_i;\mu_{c},\Sigma_{c})}
\end{equation}

Applying the classification loss only cannot reach our goal of forcing the latent space to be a Gaussian mixture distribution. There will be situations where a $z_i$ can be far away from the corresponding target class centroid $\mu _c$ and still be correctly classified since it is relatively closer to $\mu _c$ than to the means of the other classes in multiple classifications--which could be an outlier. To fix this problem, we then use a likelihood to measure the extent to of the training data fits the Gaussian mixture distribution. The \textbf{likelihood} for $\{z, c\}$ is expressed in Equation \ref{eq:lkd}. The likelihood could serve as a constraint to the original classification loss.

\begin{equation} \label{eq:lkd}
\mathcal{L}_{lkd} = -\sum_{i=1}^{N}\text{log }\mathcal{N}(z_i;\mu_{{z}_{i}},\Sigma_{{z_{i}}})
\end{equation}

Gaussian mixture loss $\mathcal{L}_{GM}$ we optimize to update the parameters of Encoder, $\mu_{c}$ and $\Sigma_{c}$, during Step 1, is defined in Equation \ref{eq:gm}, in which $\lambda$ is a weighting coefficient.

\begin{equation} \label{eq:gm}
\mathcal{L}_{GM}=\mathcal{L}_{cls}+\lambda_{lkd}\mathcal{L}_{lkd}
\end{equation}

\textbf{Label Irrelevant Branch: Encoder$_u$ and Adversarial Classifier}
We notice that generative models usually try to generate various unseen samples. However, the generation of CEs needs to satisfy different criteria, as mentioned in Section \ref{sec:intro}. In our problem setting, to satisfy \emph{Desideratum 1}, which requires maintaining the characteristics of the query sample during the search, it is intuitive to adopt disentanglement methods in the latent space of an autoencoder.

Inspired by a Two-Step Disentanglement Method \cite{hadad2020}, we introduce an Adversarial Classifier to ensure that the embedding $z_u$ captured by the Encoder\textsubscript{u} is classification label-irrelevant. 
It is inspired by GANs, where the discriminator gradually loses the capacity to tell the generated data from the real data during the training phase. 
While GANs are usually used to improve the quality of generated output -- telling fake from real, the adversarial component in our design encourages the Encoder\textsubscript{u} to dismiss information about the labels, leading to disentanglement. With the Adversarial Classifier, we could guarantee that the $z_u$ and $\hat{y}$ are disentangled and independent, which prepares for the interpolation in the Generating Step. $\hat{y}'$ is the classification label of $z_u$ through the Adversarial Classifier. The \textbf{adversarial classification loss} is shown in Equation \ref{eq:acl} as binary cross entropy loss.

\begin{equation} \label{eq:acl}
\mathcal{L}_{adv} = -\frac{1}{N}\sum_{i=1}^{N}\hat{y_i}\text{log}\hat{y_i}'+(1-\hat{y_i})\text{log}(1-\hat{y_i}')
\end{equation}

\textbf{Autoencoder} 
The reconstruction error of the autoencoder is shown in Equation \ref{eq:rec}. 

\begin{equation} \label{eq:rec}
\mathcal{L}_{rec}=\frac{1}{N}\sum_{i=1}^{N}||x_i-x'_i||^2
\end{equation}
\textbf{Summary} The configuration of the network in the Training Step is composed of three network branches: first, in the \textbf{Label Relevant Branch}, the $\mathcal{L}_{GM}$ forces $z$ to be Gaussian Mixture Distribution. Second,  in the \textbf{Label Irrelevant Branch}, the Adversarial Classifier is trained to minimize the \textbf{adversarial classification loss} $\mathcal{L}_{adv}$ in Equation \ref{eq:acl} -- it is trained to classify $z_u$ to $\hat{y}$. Third, the autoencoder network is trained to minimize the \textbf{total loss} $\mathcal{L}$, the sum of three terms as shown in Equation \ref{eq:recon}: (i) the \textbf{reconstruction loss} $\mathcal{L}_{rec}$ as shown in Equation \ref{eq:rec}, (ii) \textbf{likelihood} in Equation \ref{eq:lkd} and (iii) minus the \textbf{adversarial classification loss} $\mathcal{L}_{adv}$ in Equation \ref{eq:acl}. 
\begin{equation} \label{eq:recon}
\mathcal{L} = \mathcal{L}_{rec}+\lambda_{lkd} \mathcal{L}_{lkd}-\lambda_{adv} \mathcal{L}_{adv}
\end{equation}


\subsection{Algorithm} \label{sec:algorithm}
Algorithm \ref{alg:main} and \ref{alg:main2}  show pseudo code to process our method in addition to Figure \ref{fig:1} which is used to generate the counterfactual in Section \ref{sec:eva}. 

\begin{algorithm}[bt]
\caption{Training Step of our proposed architecture }
   \begin{algorithmic}[1]
   
   \Require $\psi$, $\pi$, $\phi$ and $\omega$ the initial parameters of Encoder, Encoder$_u$,  Decoder and Adversarial Classifier; $\mu_{c}$ and $\Sigma_{c}$ the initial mean and covariance of the Gaussian distribution of $z$; $n$  the number of iterations; $\lambda_{lkd}$ and $\lambda_{adv}$ the weights of regularization terms $\mathcal{L}_{lkd}$ and $\mathcal{L}_{adv}$; $c\in [0,1]$
 
  \While {not converged} 
    \For {$i=0$ to $n$}
      \State Sample $\{x, y\}$ a batch from dataset $\mathcal{D}_t$.
          \State $\omega \stackrel{+}{\gets} -\bigtriangledown_\omega \mathcal{L}_{adv}$
      \State $\mathcal{L}_{GM} \gets \mathcal{L}_{cls}+\lambda_{lkd} \mathcal{L}_{lkd}$
     \State $\psi, \mu_{c}, \Sigma_{c}\stackrel{+}{\gets}-\bigtriangledown_{\psi, \mu_{c}, \sigma_{c}} (\mathcal{L}_{GM}-\lambda_{adv}\mathcal{L}_{adv})$
     \State Sample $\{x, y\}$ a batch from dataset $\mathcal{D}_t$.
     \State $\mathcal{L} \gets \mathcal{L}_{rec}+\lambda_{lkd} \mathcal{L}_{lkd}-\lambda_{adv}\mathcal{L}_{adv}$
     \State $\psi,\pi,\phi \stackrel{+}{\gets} -\bigtriangledown_{\psi,\pi,\phi} \mathcal{L}$
     \EndFor 
  \EndWhile

   \end{algorithmic}
   \label{alg:main}
   \end{algorithm}

\begin{algorithm}[bt]

\caption{Generating Step of our proposed architecture}
   \begin{algorithmic}[1]

  \Require $S$ samples $\alpha=\alpha_{s=1}^S$ in $(0,1]$; $x_q$ the query sample;  $x_s$ potential CE;  $f$ classifier; $tol$ tolerance; $T$ probability of target counterfactual class ($0.5$ for decision boundary); $\mu_1$ the mean of target class in latent space of the label relevant branch
  \State $z_q \gets Encoder(x_q)$
  \State $z_{u,q} \gets Encoder_u(x_q)$
    \For {$\alpha_s$ in $\alpha$}
    \State $z_s = (1- \alpha)z_q + \alpha \mu_1$
    \State $x_s \gets Decoder(z_s,z_{u,q})$
    \If {$|f(x_s)-T|< tol$ }
     \State $x_{cf}=x_s$
     \EndIf
   \EndFor
   \State\Return {$x_{cf}$}
   
   \end{algorithmic}
   \label{alg:main2}
   \end{algorithm}

\begin{figure*}[bt]
    \centering  \includegraphics[width=0.8\textwidth]{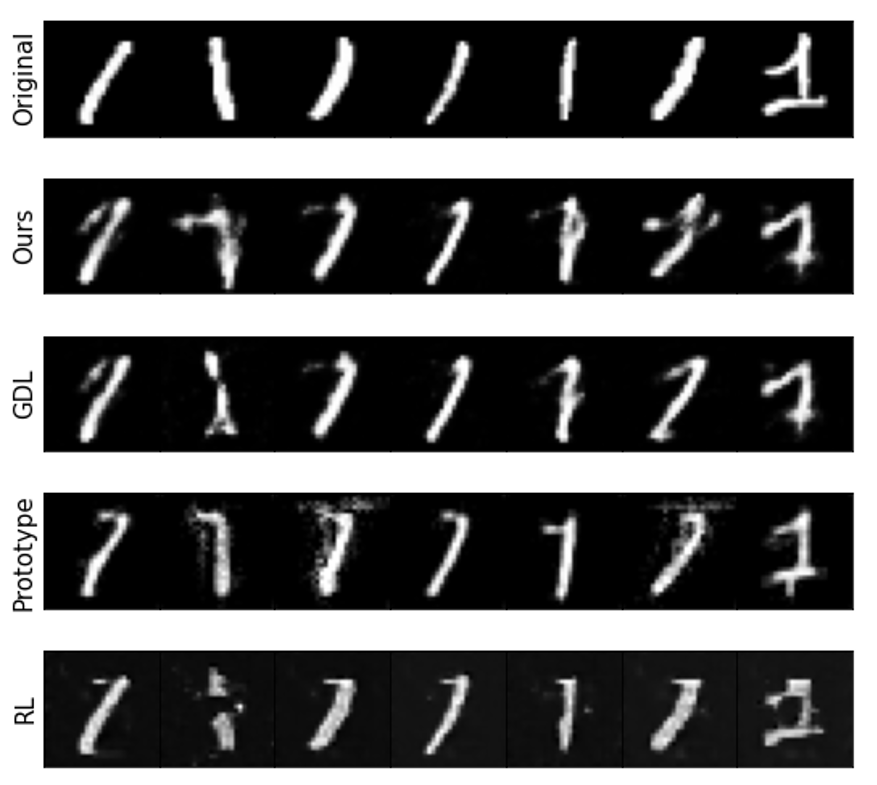}
    \caption{Original and counterfactual samples for MNIST showing from top to bottom: original query
images and CEs found via Our Method, GDL, Prototype and RL. Note Our Method exhibits more combination of the characteristics of the original query samples (e.g., tilt and thickness of the strokes) and the target label than the other methods. Our Method and GDL remain (perturbation in latent space) more realistic than the Prototype and RL (perturbation in original space). 
    }
    \label{fig:comparison}
\end{figure*}

For the Training Step, we sample a batch from the training dataset to update the parameters of Adversarial Classifier $\omega$, Encoder $\psi$ and mean and covariance $\mu_b$, $\mu_t$, and $\sigma_b$, $\sigma_t$ -- base class and target class combined as training data. Then we sample another batch to update the Encoder, Encoder$_u$, and Decoder parameters together($\psi$, $\pi$, and $\phi$). We iterate this procedure until convergence of the loss functions is reached. During the end of the Training Step: we should get a \textbf{label relevant} latent space $z$ following Gaussian Mixture distribution (see Figure \ref{fig:pca} for PCA of $z$ in Appendix) and a \textbf{label irrelevant} latent space $z_u$ to capture the label irrelevant information of samples.  
In Generating Step, where we generate a CE of a query sample $x_q$, We first pass $x_q$ from the base class through the Encoder and the Encoder$_u$ to obtain the sample’s embeddings $z_q$ and $z_{u,q}$. Then we get the searched potential $z_s$ by linear interpolation in latent space: $z_s = (1- \alpha)z_q + \alpha \mu_1$ ( $\alpha$ in $(0, 1]$ ), where $\mu_1$ is the mean of target class in latent space of the label relevant branch. $z_s$ and $z_{u,q}$ combined are decoded through the Decoder to get $x_s$, and the pre-trained classifier $f$ assesses its classification score. The counterfactual search stops if it crosses the decision boundary and is within a user (end-user or developer of the algorithm) specified tolerance $tol$, which is the desired maximum distance from $T$ for the generated counterfactual’s classification score. For the experiments in Section \ref{sec:eva}, the decision boundary $T$ is set to be 0.5. The search is performed by sampling along the line with a finite number of $\alpha$. 

Note that we demonstrate our method in the scenario with binary decision. However, it is straight-forward to extend our method to handle a multi-categorical decision. 
\begin{table*}[bt]
\centering
\caption{\label{tab:mnist}Summary of metrics for MNIST}
\begin{tabular}{|l|l|l|l|l|l|}
\hline
Method     & time(s)           & reconstruction           & sparsity   & validity(\%) & proximity          \\ \hline
Our Method & \textbf{0.007$\pm$0.001} & \textbf{0.012$\pm$0.003} & 2.119 $\pm$0.857        & \textbf{89.5 }    & 0.193  $\pm$0.075       \\
GDL        & 1.568 $\pm$0.047         & 0.247 $\pm$0.048        & 2.121 $\pm$0.085       & 77.9    & \textbf{0.172$\pm$0.068 }         \\
Prototype        & 1.345 $\pm$0.095         & 0.926   $\pm$0.175       & \textbf{0.014$\pm$0.005} & 58.7   & 1.075$\pm$0.055 \\
RL        & 0.085   $\pm$0.012      & 0.857$\pm$0.058         & 0.015$\pm$0.026 & 60.5  & 0.834$\pm$0.047 \\ \hline
\end{tabular}
\end{table*}

\begin{table*}[bt]
\centering
\caption{\label{tab:adult}Summary of metrics for Adult income}
\begin{tabular}{|l|l|l|l|l|l|}
\hline
Method     & time(s)            & reconstruction           & sparsity   & validity(\%) & proximity           \\ \hline
Our Method & \textbf{0.008$\pm$0.002} &  \textbf{0.012$\pm$0.005}          & 0.343$\pm$0.127          & \textbf{90.3}   & \textbf{0.193$\pm$0.045}          \\
GDL        & 1.568$\pm$0.224       & 0.520$\pm$0.135 & 0.090$\pm$0.002          & 84.2   & 2.170$\pm$0.835          \\
Prototype        & 7.345$\pm$0.784         & 4.463$\pm$0.563          & 0.014$\pm$0.005 & 75.3   & 1.075$\pm$0.235 \\
RL        & 1.804$\pm$0.112       & 5.453$\pm$0.673         & \textbf{0.012$\pm$0.008} & 65.3  & 0.875$\pm$0.132 \\
\hline
\end{tabular}
\end{table*}

\section{Experiments and Evaluation} \label{sec:eva}

We compare our method to three other counterfactual methods introduced in Section \ref{sec:back}: Gradient Descent Method improved with Prototypes (Prototype) \cite{vanlooveren2020}, CE Generation with Reinforcement Learning (RL) \cite{samoilescuModelagnosticScalableCounterfactual2021} and Gradient Descent in the Latent
Space of a VAE (GDL) \cite{balasubramanian2021} on three datasets: MNIST \cite{zotero-844} (image), Adult Dataset \cite{zotero-1218} (tabular) and Lending Club \cite{zotero-1220} (tabular). The reason why we choose Prototype and RL is that they are both designed not only for image datasets but also for tabular datasets. Besides, they both use autoencoders to learn the representation to remain close to the data distribution, similar to our design, while they use other constraints to ensure desiderata like sparsity. We use the package ALIBI \cite{klaiseAlibiExplainAlgorithms2021} for Prototype and RL to stay close to the original design. GDL is a relatively simple baseline, but it is similar to our method because both operate in latent space. We implement GDL from scratch since the source code is not available. Our networks are trained on an Intel(r) Core(TM) i7-8700 CPU and the networks in our experiments are built based on Pytorch \cite{NEURIPS2019_9015}

\subsection{Measures for Comparison} We evaluate the quality of CEs by the measures taken from literature \cite{verma2020,barr2021,wachterCounterfactualExplanationsOpening2018a}:  (i) \textbf{counterfactual generation time}, time required to find a CE for a given query sample (ii) \textbf{validity}, the
percentage success in generating CEs that the target labels requested by the users are reached (iii) \textbf{proximity}, distance($L_2$
norm) from query samples to CEs in original space (iv) \textbf{sparsity}, $L_1$
norm of the change vector in original space (v) \textbf{reconstruction loss} -- a measure of the CE being in sample. We pass a CE through the autoencoder and measure its loss. A smaller loss indicates closer to the original data distribution because the autoencoder is trained on the same training dataset as the pre-trained classifier. Further details about the metrics used in the experiments can be found in the appendix.

\subsection{Datasets, Training and Evaluation}

\begin{figure*}[bt]
   
    \centering
    \includegraphics[width=13cm]{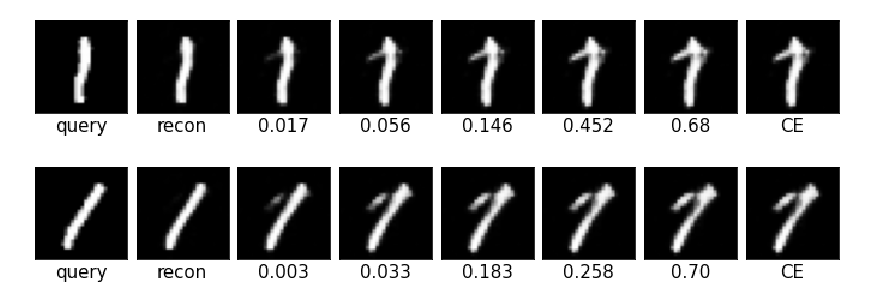}
    \caption{the gradual changes and classification scores for query samples along the interpolation path. Along the paths, the label irrelevant characteristics (e.g., tilt and thickness of the strokes) stay.}
    \label{fig:grad}
\end{figure*}

{\bf{MNIST}} The MNIST database is a large database of handwritten digits. For MNIST, we adjust the problem as a binary classification problem of predicting ones (our base class) and sevens (our target class) while the original problem setting is multi-calcification. MNIST provides a naturally intuitive visualization of the result of a gradual change from the base class to the target class crossing the decision boundary given a query sample. The counterfactual generated shown in Figure \ref{fig:grad} exhibits a combination of characteristics from the query sample (e.g., the tilt of long-stroke) and from the target class (e.g., longer leveled stroke in sevens). Our \emph{Desideratum 1} -- generating a CE  while keeping the characteristics of the query sample which are not related to the classification prediction is reached. In our problem setting with MNIST, the tilt and length of long strokes of query sample ones are kept during the interpolation process. Table \ref{tab:mnist} (mean±SD) shows the quality of counterfactual explanations as measured by the above metrics. Our method outperforms the other methods in time, reconstruction, and validity but not in proximity and sparsity (very close to GDL in proximity, though). Our method outperforms Prototype and RL in the time dimension since it searches through the latent space with lower dimensions. It is also more substantial than GDL because it performs interpolation instead of optimization for a single query sample. It indicates that our method is suitable for high-dimensional applications which require intensive computation. 

\begin{table*}[bt]
\caption{\label{tab:lc}Summary of metrics for Lendidng Club default loan}
\centering
\begin{tabular}{|l|l|l|l|l|l|}
\hline
Method     & time(s)           & reconstruction           & sparsity   & validity(\%) & proximity          \\ \hline
Our Method & \textbf{0.007$\pm$0.001} &  \textbf{0.132$\pm$0.081}          & 1.713$\pm$0.627          & \textbf{92.1}   & 3.532$\pm$0.258          \\
GDL        & 3.569 $\pm$0.087         & 0.142 $\pm$0.046        & 1.731 $\pm$0.068       & 76.3    & 1.652$\pm$0.668         \\
Prototype        & 3.137 $\pm$0.915         & 0.546   $\pm$0.235       & \textbf{1.016$\pm$0.035} & 75.7   & 1.975$\pm$1.045 \\
RL        & 0.035   $\pm$0.042      & 0.673$\pm$0.124         & 1.017$\pm$0.076 & 67.5  & \textbf{1.034$\pm$0.127}  \\ \hline
\end{tabular}
\end{table*}

For MNIST, we use a CNN-based network, and we train the model for 20 epochs by stochastic gradient descent using the Adam optimizer and a batch size of 100 on the training dataset. Our model uses two hyperparameters, $\lambda_{adv}$ and $\lambda_{lkd}$, for regularizing the loss functions for the network. We varied $\lambda_{adv}$ and $\lambda_{lkd}$ between 0.01 and 1 during training. During the training process, we want the three loss terms $\mathcal{L}_{adv}$, $\mathcal{L}_{lkd}$ and $\mathcal{L}_{cls}$ to remain approximately on the same scale. We found $\lambda_{adv}= 0.05$ and $\lambda_{lkd}= 0.1$ to give us a numerical balance among the three loss terms by checking the testing dataset. We use these values to report our results. The details of the hyper-parameters of the Discriminator are shown in the Appendix. 

{\bf{ADULT}} The Adult dataset was drawn from the 1994 United States Census Bureau data. It used personal information such as education level and working hours per week to predict whether an individual earns more or less than \$50,000 per year \cite{kohavi1996}. We train a classifier with age, years of education, capital gain, capital loss, hours-per-week as continuous features, and education level as a categorical feature for simplicity (considering mutable and immutable features are not the focus of this paper). The dataset is imbalanced -- the instances made less than \$50,000 constitute 25\% of the dataset, and the instances made more than \$50,000 constitute 75\% of the dataset. To avoid the situation that the accuracy of the classifier for an imbalanced dataset only reflects the distribution of the training dataset, we train a classifier with re-weighting according to the proportion of base and target class. For tabular datasets, more prepossessing is performed compared to image datasets. We normalize the continuous features and use one-hot encoding to deal with the categorical features for the input of the autoencoder. We train the model for 100 epochs by stochastic gradient descent using the Adam optimizer and a batch size of 100. $\lambda_{adv}$ and $\lambda_{lkd}$ are set at 0.05 and 0.5. For tabular datasets, we use a multi-layer perceptron-based network. The details of the architectures of the networks for the Adult income dataset are shown in the appendix. Based on the comparison results shown in Table \ref{tab:adult}, our method outperforms in time, reconstruction dimensions and proximity, while RL is stronger in sparsity. Intuitively, perturbation in original space without much guidance could quickly end up as adversarial samples \cite{goodfellow2015}, which provide non-actionable CEs. In contrast, operations in latent space (our method and GDL) could be closer to the original datasets.

{\bf{Lending Club}} Lending Club is a peer-to-peer lending company that allows individuals to lend to other individuals \cite{zotero-952}. This tabular dataset includes whether a borrower defaulted on their loan size, annual income, debt-to-income ratio, FICO score, loan length, etc. We train a classifier to predict default using five continuous and one categorical feature. Since this dataset is also imbalanced -- the default class constitutes around 15\% of the population, we adjust the loss functions of the classifier and the autoencoder to re-weighted versions based on the proportion of each class. From Table \ref{tab:lc}, we find that similar to the Adult income dataset, our method is much faster at generating CEs and excels at reconstruction and validity. The training details of this dataset are similar to ADULT. Furthermore, more details are included in the Appendix.

\section{Conclusion}
\label{sec:con}
This paper presents a novel model-agnostic algorithm for finding CEs via linear interpolation in latent space. Our method implements a framework that first disentangles the label relevant and label irrelevant dimensions and then searches in a Gaussian mixture distributed latent space of the label relevant latent dimensions for CEs given a query sample. We demonstrated our method’s advantages and disadvantages by comparing it to three similar methods (GDL, Prototype and RL) on three different datasets (MNIST, ADULT income, and Lending Club default loan). We show that our method is faster and provides more valid CEs which are closer to the original dataset (in the dimensions of time, validity and reconstruction). Based on the presented comparison, we suggest that our work could evolve around improving latent representation.

\section*{Acknowledgement}
This work has received funding from the European Union’s Horizon 2020 research and innovation programme under Marie Sklodowska-Curie Actions (grant agreement number 860630) for the project “NoBIAS - Artificial Intelligence without Bias”.

\bibliographystyle{named}
\bibliography{ijcai23}

\appendix

\subsection{Metrics for Comparison}

\begin{figure*}
\subfloat[MNIST dataset\label{fig:1c}]
  {\includegraphics[width=.325\linewidth]{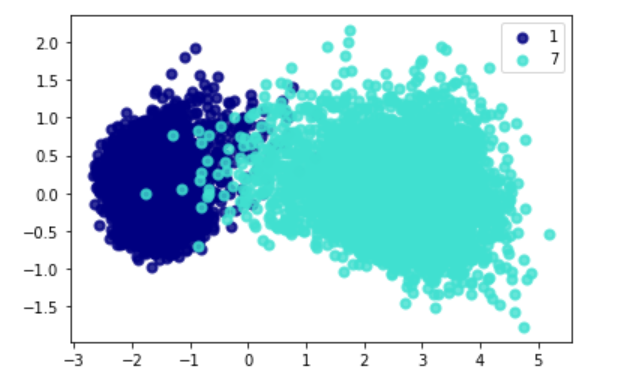}}\hfill
\subfloat[Adult dataset\label{fig:2c}]
  {\includegraphics[width=.3\linewidth]{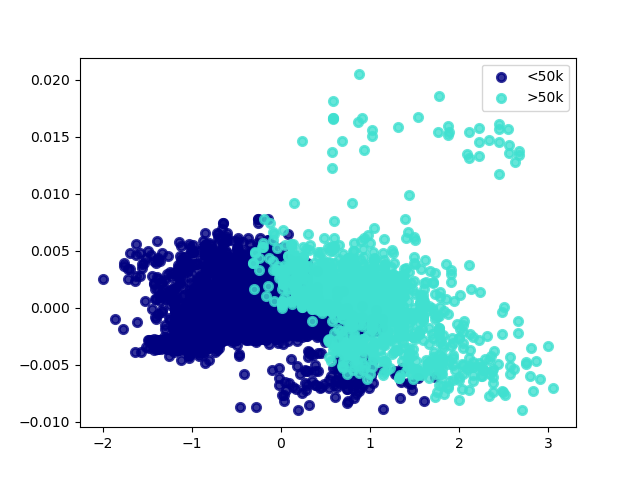}}\hfill
\subfloat[Lending Club dataset\label{fig:3c}]
  {\includegraphics[width=.3\linewidth]{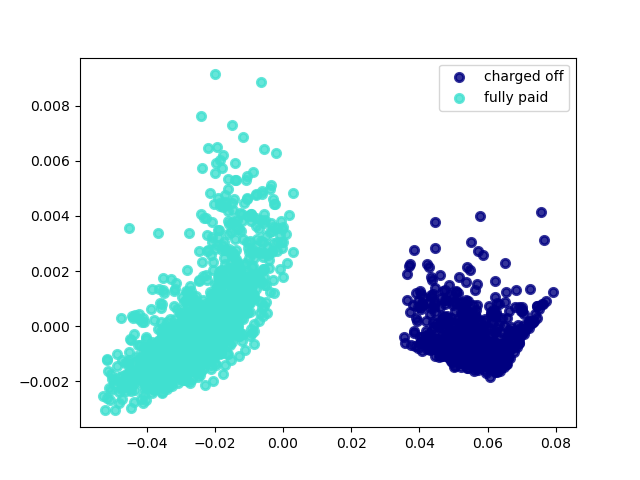}}
\caption{PCA of the base and target class in latent space} \label{fig:pca}
\end{figure*}
We evaluate the quality of counterfactuals by the metrics taken from literature: 
\begin{inparaenum}[1)]
\item time - time required to find a counterfactual for a query sample. Time is one of the most significant strengths of our method. We use the time spent to generate a counterfactual explanation for a query sample on average. Our approach is particularly fast compared to other methods because generating a CE requires a search through a relatively lower dimension and projection through the decoder for no more than a fixed number of interpolated points (search stops when criteria are met). By comparison, iterative search based on gradients or perturbing in the input space can be quite expensive if the search distance is large, the learning rate is low, or the mutable dimensions are high. We do not include the time of training the autoencoder with Gaussian mixture latent space due to the fact that this training time is not trivial. It is a one-time cost and is not related to the scalability of CEs generation for a query sample. Ideally, our method triumphs even more, when the query sample is high dimensional. 
\item validity - the percentage of success in generating counterfactuals cross the decision boundary. The goal for generating counterfactuals is that they are counterfactuals, which requires them to be classified as the target class. We evaluate validity by including the pre-trained classifier in the algorithm. Since linear search is performed with the classifier is ended when certain criteria are met for the classifier. We could see, in general, that the performance of this dimension for the three methods is satisfying since all three methods employ the classifier as the end search criteria. \item proximity - a measure of distance from query sample to counterfactual in latent space. Proximity is used as a constraint to promote the performance of counterfactual generation \cite{verma2020}. This metric measures how close a counterfactual is to the query sample it is generated from. Usually, this measure is calculated by the $L2$ norm in the input space. \item sparsity - a measure of features changed. Sparsity is a common metric in the counterfactual literature \cite{verma2020}. Sparsity is a measure of how sparse the change vector is. We calculate the $L1$ norm of the change vector in input space. \item reconstruction loss - a measure of the CE being close to data manifold \cite{barr2021}. We can measure the closeness of a CE to its original dataset by passing a counterfactual through the autoencoder and measuring its reconstruction loss. During the training process of the autoencoder, we try to minimize the reconstruction loss. Intuitively, a sample with a smaller loss should be more in-sample and closer to the original data distribution. An unseen sample should have a larger reconstruction loss concerning the autoencoder. 
\end{inparaenum} 
\subsection{PCA of latent space}

\subsection{Further Thoughts about the Comparison Results}
We could see that our method performs better on three dimensions above: counterfactual generation time, validity and reconstruction loss due to its design. We design the architecture based on our three desiderata and in the end, through the experiment, we show that the desiderata are achieved. Our method performs linear interpolation in the latent space instead of optimization in the original space, it has the highest validity, and also because the linear interpolation happens in latent space, we cannot guarantee the sparsity and proximity in the original space as shown in the tables, which is stated in Section \ref{sec:back}.

\subsection{Algorithms of GDL}

In GDL, instead of updating parameters of DNN which is commonly applied, we update the input query sample and the latent projection of the query sample correspondingly until it cross the decision boundary and the tolerance is satisfied.

\begin{algorithm}
\caption{Gradient Descent Latent Space}
   \begin{algorithmic}[1]
   \label{alg:gdl}
   \Require $\psi$ and $\phi$  the initial parameters of Encoder and Decoder; $f$ classifier; $x_b$ the query sample; $tol$ tolerance; $p$ probability of target counterfactual class ($0.5$ for decision boundary)
  
      \State $z_b \gets \psi(x)$
      \State $z_t \gets z_b $
      
      \State $z_t \stackrel{+}{\gets} \bigtriangledown_{z_{t}} (T -f(\phi(z)))^2$
      \State $x_t \gets \phi(z_t)$

    \If {$|f(x_t)-T|< tol$ \bf{and} $f(x_t)>p$}

   \State $x_{cf}=\phi(z_t)$
   \State\Return $x_{cf}$
   \EndIf
   \end{algorithmic}
   \end{algorithm}

\subsection{Repetition} We repeat experiments on each dataset five times. Before each repetition, we randomly split data into training data (80\%) and test data (20\%) for the computation of the standard errors of the metrics.

\subsection{Details of Autoencoders and Classifer}

We tabulate the detailed structures of autoencoders and classifiers during experiment in Section \ref{sec:eva}. For tabular datasets, to avoid break the differentiation of $(T -f(\phi(z)))^2$, instead of argmax for decoder output, we apply temperature annealing softmax to have a sharper reconstructed sample where we set $t=0.5$. 

\begin{table}[H]
\centering
\caption{\label{tab:adultc_1} The network structure of MNIST
classifier}
\begin{tabular}{|l|}
\hline
Input 28*28            \\ \hline
Linear 28*28-128, ReLu \\ \hline
Linear 128-64, ReLu    \\ \hline
Linear 64-32, ReLu     \\ \hline
Linear 32-1, Sigmoid   \\ \hline
\end{tabular}
\end{table}

\begin{table}[H]
\centering
\caption{\label{tab:adultautoencoder} The network structure of Adult income autoencoder}
\begin{tabular}{|l|l|}
\hline
Encoder                                               & Decoder                                                       \\ \hline
Input(1,28,28)                                        & Linear(25,128), Relu                                    \\ \hline
Conv2d(1,8,2), stride 1,  ReLu                  & Linear(128, 29*29*32), unflatten                    \\ \hline
Conv2d(8,16,2), stride 1,  BatchNorm2d 16,ReLu & convtranspose2d(32,16,2),stride 1, batchnorm2d 16,  ReLu \\ \hline
Conv2d(16,32,2), stride 1, Leaky ReLu, flatten        & convtranspose2d(16,8,2),stride 1, batchnorm2d 8, ReLu   \\ \hline
Linear(22912, 128), ReLu,                       & convtranspose2d(8,1,2), stride 1, sigmoid                     \\ \hline
Linear(128,15)                                        & output(1,28,28)                                               \\ \hline
\end{tabular}
\end{table}

The structure of Encoder$_u$ is similar to Encoder but with the last layer of 25 nodes.
\begin{table}[H]
\centering
\caption{\label{tab:adultc} The network structure of Adult income classifier}
\begin{tabular}{|l|}
\hline
Input $x$(cont\_in 6 + cat\_in 16/one-hot encoding) \\ \hline
Linear 21-10, ReLu                                \\ \hline
Linear 10-4, ReLu                                 \\ \hline
Linear 4-2, ReLu                                  \\ \hline
Linear 2-1, Sigmoid                               \\ \hline
\end{tabular}
\end{table}

\begin{table}[H]
\centering

\caption{\label{tab:adultautoencoder_2} The network structure of Adult income autoencoder}

\begin{tabular}{|l|ll|}
\hline
Encoder                                                   & Decoder                                &                                                  \\ \hline
Input x(cont\_in 6/min-max + cat\_in 16/one-hot encoding) & Linear 3-6, LeakyReLu                  &                                                  \\ \hline
Linear 21-12, LeakyReLu                                   & Linear 6-12, LeakyReLu                 &                                                  \\ \hline
Linear 12-24, LeakyRelu                                   & Linear 12-24, LeakyReLu                &                                                  \\ \hline
Linear 24-12, LeakyReLu                                   & \multicolumn{1}{l|}{Linear 24-5, Tanh} & Linear 24-16, softmax \\ \hline
Linear 12-6, LeakyReLu                                    & \multicolumn{1}{l|}{cont\_output 5}    & cat\_output 16                                   \\ \hline
Linear 6-3                                                & \multicolumn{1}{l|}{}                  &                                                  \\ \hline
\end{tabular}
\end{table}


\end{document}